\documentclass[10pt,twocolumn,letterpaper]{article}

\usepackage{subcaption}
\usepackage{multicol}
\usepackage{misc/cvpr}
\usepackage{times}
\usepackage{amsmath}
\usepackage{amssymb}
\usepackage{algorithm}
\usepackage{algpseudocode}
\usepackage{xspace}
\newcounter{savefootnote}
\usepackage{capt-of}
\usepackage{graphicx}

\usepackage{longtable}

\usepackage[pagebackref=true,breaklinks=true,colorlinks,bookmarks=false,colorlinks = true,
            linkcolor = blue,
            urlcolor  = blue,
            citecolor = blue,
            anchorcolor = blue]{hyperref}

 \cvprfinalcopy 
\newcommand\enote[1]{{\textcolor{blue}{\textbf{Elad}: #1}}}
\renewcommand\enote[1]{{\textcolor{blue}{}}}

\def\secspace{\vspace{-0.2cm}}
\newcommand\ignore[1]{{}}

\newcommand{\argmin}[1]{\underset{#1}{\operatorname{argmin}}}

\def\calF{\mathcal{F}}

\def\calL{\mathcal{L}}
\def\calR{\mathcal{R}}
\def\calG{\mathcal{G}}

\def\indepth{I}
\def\outdepth{O}
\def\inwidth{w}
\def\inheight{x}
\def\outwidth{y}
\def\outheight{z}
\def\filtwidth{f}
\def\filtheight{g}
\def\inactive{A}
\def\outactive{B}

\def\etc{{\em etc.}\xspace}
\def\eg{{\em e.g.,}\xspace}
\def\ie{{\em i.e.,}\xspace}
\def\aka{{\em a.k.a.}\xspace}

\newcommand{\figref}[1]{Figure~\ref{#1}}

\newcommand{\secref}[1]{Section~\ref{#1}}

\def\DNN{\text{DNN}\xspace}
\def\DNNs{{\DNN}s\xspace}
\def\FLOP{{\text{FLOP}\xspace}}
\def\FLOPs{\text{FLOP}s\xspace}
\def\resnet{ResNet101\xspace}
\def\mobilenet{MobileNet\xspace}

\def\FlexNet{MorphNet\xspace}

\def\naive{na{\"i}ve\xspace}

\def\cvprPaperID{2508} 

\def\citep{\cite}
\def\citet{\cite}

\newcommand{\comment}[1]{}
\newcommand{\eq}[1]{Eq.~(\ref{#1})}

\ifcvprfinal\fi

\makeatletter
\renewcommand*{\@fnsymbol}[1]{\ensuremath{\ifcase#1\or *\or \dagger\or \ddagger\or
    \mathsection\or *\or \|\or **\or \dagger\dagger
    \or \ddagger\ddagger \or 1 \else\@ctrerr\fi}}
\makeatother

 \title{
MorphNet: Fast \& Simple Resource-Constrained Structure Learning of Deep Networks}

\author{Ariel Gordon\thanks{Google Research}\\
{\tt\small gariel@google.com}
\and
Elad Eban\footnotemark[1]\\
{\tt\small elade@google.com}
\and
Ofir Nachum\thanks{Google Brain}\\
{\tt\small ofirnachum@google.com}
\and
Bo Chen\footnotemark[1]\\
{\tt\small bochen@google.com}
\and
Hao Wu\footnotemark[1]\\
{\tt\small haou@google.com}
\and 
Tien-Ju Yang
{\footnotemark[1]~\thanks{Energy-efficient multimedia systems group, MIT.
}}
\\
{\tt\small tjy@mit.edu }
\and
Edward Choi
\footnotemark[1]~\thanks{Georgia Institute of Technology.}  \\
{\tt\small mp2893@gatech.edu}
}

\begin{document}
\setcounter{savefootnote}{\value{footnote}}
\setcounter{footnote}{0}
\renewcommand{\thefootnote}{\alph{footnote}}

\maketitle
 \begin{abstract}
We present \FlexNet, an approach to automate the design of neural network structures. \FlexNet iteratively shrinks and expands a network,
shrinking via a resource-weighted sparsifying regularizer on activations and expanding via a uniform multiplicative factor on all layers. 
In contrast to previous approaches, our method is scalable to large networks, adaptable to specific resource constraints (e.g. the number of floating-point operations per inference), and capable of increasing the network's performance. When applied to standard network architectures on a wide variety of datasets, our approach discovers novel structures in each domain, 
obtaining higher performance while respecting the resource constraint.
\end{abstract}

 \secspace\section{Introduction}

The design of deep neural networks (\DNNs) has often been more of an art than a science. Over multiple years, top world experts have incrementally improved the accuracies and speed at which \DNNs perform their tasks, harnessing their creativity, intuition, experience, and above all - trial-and-error.  
Structure design in \DNNs has thus become the new feature engineering. 
Automating this process is an active research
field that is gaining significance as \DNNs 
become more ubiquitous in a variety
of applications and platforms.

One key approach towards automated architecture search 
involves sparsifying regularizers. 
Initially it was shown that applying L1 regularization on weight matrices can reduce the number of nonzero weights with little effect on the performance (e.g. accuracy or mean-average-precision)
of the \DNN~\cite{williams1995bayesian, collins2014memory}. 
However, as \DNNs started powering more and more industrial applications, practical constraints such as inference speed and power consumption became of increasing importance. 
Standard L1 regularization can prune individual
connections (edges) in a neural network, but
this form of sparsity is ill-suited to 
modern hardware accelerators and does not result
in a speedup in practice.
To induce better sparsification,
more recent work has designed regularizers 
which target neurons (\aka activations) rather than 
weights~\cite{lebedev2016fast,wen2016learning,alvarez2016learning}.  While these techniques have succeeded in reducing
the number of parameters of a network, they do not target reduction of a \emph{particular} resource (\eg the number of floating point operations, or \FLOPs, per inference). In fact, resource specificity of sparsifying regularizers remains an under explored area. 

A more recent approach to
neural network architecture design expands the scope
of the problem from only shrinking a network to 
optimizing every aspect of the network structure.
Works using this approach~~\citep{zoph,supernets,real2017large,nemo} 
rely on an auxiliary neural network to learn the art of neural network design from a large number of trial-and-error attempts.  While these proposals have succeeded in achieving new state-of-the-art results on several datasets~\cite{real2017large,zoph2016neural}, they have done so at the cost of an exorbitant number of trial-and-error attempts. These methods require months or years of GPU time to obtain a single architecture, and become prohibitively expensive as the networks and datasets grow in complexity and volume.

\begin{figure*}[t]
\vspace{-.3cm}\begin{center}
   \includegraphics[width=0.99\linewidth]{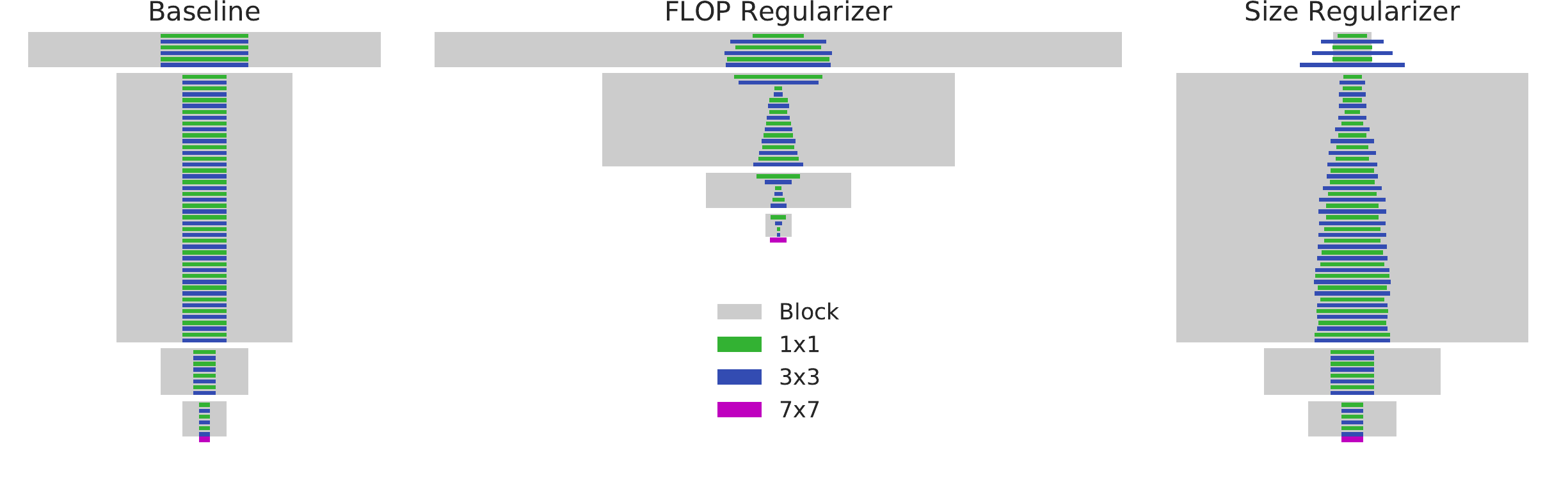}
\end{center}
    \vspace*{-6mm}
    \caption{
    {\small \resnet based models with similar performance (around $0.426$ MAP on JFT, see \secref{sec:empirical}). A structure obtained by shrinking \resnet uniformly by a $\omega=0.5$ factor (left), and structures learned by \FlexNet when 
    targeting \FLOPs (center) or model size (\ie number of parameters; right).
    Rectangle width is proportional to the number of channels in the layer and residual blocks are denoted in gray.
    $7\times7$, $3\times3$, and $1\times1$ convolutions are in purple, blue and green respectively. 
    The purple bar at the bottom of each model is thus the input layer. 
    Learned structures are markedly different from the human-designed model and from each other. 
    The \FLOP\, regularizer primarily
    prunes the early, compute-heavy layers. 
    It notably learns to \textit{remove whole layers} to further reduce computational burden. 
    By contrast, the model size regularizer 
    focuses on removal of $3\times3$ convolutions 
    at the top layers as those are the most parameter-heavy.}
    }
\label{fig:structures}
\end{figure*}

Given these various research directions, automatic neural network architecture design is currently effective only under limited conditions and given knowledge of the right tool to use.
In this paper, we hope to alleviate this issue. We 
present \FlexNet, a simple and general technique for resource-constrained optimization of \DNN architectures.




Our technique has three advantages:
(1) it is scalable to large models 
and large datasets;
(2) it can optimize a \DNN structure targeting a specific resource, such as \FLOPs per inference, while allowing the usage of untargeted resources, such as model size (number of parameters), to grow as needed;
 (3) it can learn a structure that \emph{improves} performance while reducing the targeted resource usage.

We show the efficacy of \FlexNet 
 on a variety of datasets.
As a testament to its scalability,
we find that on the JFT dataset~\citep{distill},
a dataset of 350M images and 20K classes,
our method achieves 2.1\% 
improvement in evaluation MAP while maintaining the same number of \FLOPs per inference.
The resources required by our technique to achieve this improvement are \emph{only slightly greater} than the resources required to train the model \emph{once}.

As evidence of our method's ability to
learn network architecture, we 
show that on Inception-v2~\cite{inceptionv2},
a network structure which has been 
hand-tuned by experts, our method finds
an improved network architecture which
leads to an increase of $1.1\%$ test accuracy on ImageNet,
again maintaining the same number of FLOPs per instance.

Lastly, to show constraint targeting, we 
present the results of applying our 
technique to a number of additional 
datasets while targeting different
constraints.  Our method is able
to find unique, improved structures
for each constraint, showing the
benefits of
constraint-specific targeting (see~\figref{fig:structures}). 

Overall, we find our method provides
a much needed general, automated, and scalable solution to
the problem of neural architecture design,
a problem which is currently only
solved by a combination of 
context-specific approaches
and manual labor.

 \secspace\section{Related Work}

The need for automatic procedures to
selectively remove or add
weights to a \DNN has been a topic
of research for several decades.

Optimal Brain Damage~\citep{leoptimal,hassibi1993} proposed
pruning the weights of a fully trained \DNN 
based on their contribution to the objective function.
Since the \DNN is fully trained, the contribution
of each parameter may be approximated using
the Hessian.
In this and similar pruning algorithms, it is often beneficial
to add a penalty term to the loss to encourage
less necessary weights to decrease in norm.
Traditionally, the penalty has taken
the form of L2 regularization, equivalent to weight-decay~\cite{denker1987large}.
Later work~\cite{williams1995bayesian}
proposed to use an L1 regularization,
which is known to induce sparsity~\cite{lasso, l2},
thus alleviating the need for sophisticated
estimates of a parameter's contribution to the loss.
We use an L1 regularization in our method for the same reasons.

An issue common to many pruning and penalty-based 
procedures for inducing network sparsity 
is that the removal of weights after training
and the penalty during training adversely 
affects the performance of the model.
Previous work~\cite{han2015learning} has noted the benefits of
a multi-step training process,
first training to induce sparsity
and subsequently training again
using the newer structure.
We utilize the same paradigm in our approach,
also finding that training a newer structure
from scratch benefits overall performance.

In this work we note that sparsity in \DNNs is 
useful only when it
corresponds to the removal of an entire neuron
rather than a single connection.  
Previous work has made this point as well.
Group LASSO~\cite{group-lasso}
was introduced to solve this problem
and has been previously applied to \DNNs~\cite{lebedev2016fast,wen2016learning,alvarez2016learning,murray2015auto}.
The specific technique we use
is based on an L1 penalty applied
to the scale variables of batch
normalization~\cite{batchnorm}.
This technique was also discovered
by a recent work~\cite{liu2017learning}
and similar ideas appear elsewhere~\cite{huang2017data}.
However, these works do not target a specific resource or demonstrate any improvement in performance. Moreover, they largely 
neglect to compare to \naive \DNN shrinking strategies, such as applying a uniform multiplier to all layer sizes, which is crucial given that they often study \DNNs that are 
significantly over-parameterized.

Previous works on sparsifying \DNNs
have traditionally focused on reducing model
size 
(\ie each individual parameter is
equally valuable)~\cite{collins2014memory, liu2015sparse, zhou2016less}.
Recent years have revealed that more nuanced 
prioritization is needed.
For example in mobile applications~\cite{howard2017mobilenets},
reducing latency is also important.
Our work is formulated in a general way, thus making it applicable
to a wide variety of application-specific
constraints. 
Our evaluation
studies model size and FLOPs-based constraints.
FLOPs-based constraints have been studied
previously~\cite{molchanov2016pruning,supernets,nemo},
although we believe our work is the first to
tackle the issue via cleverly designed
sparsifying regularizers.

Many of these previous works
focus on reducing the size of a network using sparsification. Our work supersedes 
such research,
going further to show that one may maintain the size (or FLOPs per inference) and gain an increase in performance by changing the structure of a neural network.
Other methods to learn the structure
of a neural network have been proposed, especially focusing on when and how to expand the size of a neural network~\cite{feng2015learning,chen2015net2net}.
While these techniques may be incorporated in our method,
we believe the simplicity of our
proposed iterative process
is important.  
Our method is easy to implement
and thus quick to try.

Finally, our work is distinct
from a school of methods that 
learn the network structure from a
large amount of trial-and-error attempts.
These methods use RL~\cite{zoph,supernets}
or genetic algorithms~\cite{real2017large,nemo}
with the purpose of finding a network architecture
which maximizes performance.
We note that some of these works
have begun to investigate resource-aware
optimization rather than maximizing
performance at all costs~\citep{supernets,nemo,nasnet}.
Still, the amount of computation necessary 
for these techniques makes them 
unfeasible on large datasets
and large models.
In contrast,
our approach is extremely scalable,
requiring only a small constant number
(often 2)
of automated trial-and-error attempts.
 \secspace\section{Background}

In this work, we consider deep feed-forward neural networks, 
typically composed of a stack of convolutions, biases, fully-connected layers, and various pooling layers,
and in which the output is a vector of scores.
In the case of classification, the final vector
contains one score per each class.

We number the parameterized layers of the \DNN 
$L=1,\dots,M+1$.  Each layer $L$ corresponds
to a convolution or fully-connected layer
and has an {\em input width} $\indepth_L$ and
{\em output width} $\outdepth_L$ associated with it.
In the case of a convolutional layer, $\indepth_L,\outdepth_L$
correspond to the number of input and output channels,
respectively, and $\outdepth_{L-1}=\indepth_L$ for most networks without concatenating residual connections. We consider $L=M+1$ to be the last
layer of the neural network.  
Thus $\outdepth_{M+1}$ is the size of the final
output vector.

Since a fully-connected layer may be considered
as a special case of a convolution, we
will henceforth only consider convolutions.
Thus for each layer $L=1,\dots,M+1$ we
also associate input spatial dimensions 
$\inwidth_L,\inheight_L$, output spatial dimensions
$\outwidth_L,\outheight_L$, and filter dimensions
$\filtwidth_L,\filtheight_L$.
The weight matrix associated with layer $L$ thus
has dimensions $\indepth_L\times\outdepth_L\times\filtwidth_L\times\filtheight_L$ and maps a $\inwidth_L\times\inheight_L\times\indepth_L$ input to a
$\outwidth_L\times\outheight_L\times\outdepth_L$ output.

The neural network is trained
to minimize a loss:
\begin{equation}
    \min_\theta \calL(\theta),
    \label{eq:loss}
\end{equation}
where $\theta$ is the collective 
parameters of the neural network and
$\calL$ is a loss measuring a combination of 
how well the
neural network fits the data and
any additional regularization terms
(\eg\ L2 regularization on weight matrices).

\subsection{Problem Setup}
We are interested in a procedure for automatically
determining the design of a neural network
to optimize performance\footnotemark[1]
under a constraint of limiting the consumption of a 
certain resource (\eg FLOPs per inference).
In the fully general case, this would entail
determining the widths $\indepth_L,\outdepth_L$,
the filter dimensions $\filtwidth_L,\filtheight_L$,
the number of layers $M$, which layers are connected to which,
\etc
In this paper, we restrict the task of neural network design
to only optimize over the output widths $\outdepth_{1:M}$
of all layers.
Thus we assume that we have a \emph{seed} network
design $\outdepth_{1:M}^\circ$, which in
addition to
an initial set of output widths also gives
the filter dimensions, network topology, 
and other design choices 
that are treated as fixed.
In Section~\ref{sec:extensions}
we elaborate on how our method can be extended
to optimize over these additional design choices.
However, we found that restricting
the optimization to only layer widths
can be effective while maintaining simplicity.

In formal terms, 
assume we are given a seed network
design $\outdepth_{1:M}^\circ$
and that the objective in~\eq{eq:loss} is 
a suitable proxy for the performance.
Let the constraint be denoted by
$\calF(\outdepth_{1:M})\le \zeta$ for $\calF$ monotonically
increasing in each dimension.  
In this paper, $\calF$ is either the number of
FLOPs per inference or the model size 
(\ie number of parameters), although our method is generalizable to other constraints.
We would like to find the 
optimal dimensions,
\begin{equation}
    \outdepth_{1:M}^* = \argmin{\calF(\outdepth_{1:M})\le \zeta}
    \min_\theta \calL(\theta).
    \label{eq:full-goal}
\end{equation}

 \secspace\section{Method}
We motivate our approach
by first presenting a \naive solution to~\eq{eq:full-goal}:
the \emph{width multiplier}.
Let $\omega \cdot \outdepth_{1:M} = \{\lfloor \omega\outdepth_1\rfloor,\dots,\lfloor \omega \outdepth_M\rfloor \}$
for $\omega>0$.
Observe that $\omega<1$ results in a shrunk network
and $\omega>1$ results in an expanded network.
The width multiplier (with $\omega <1$) was first introduced 
in the context of MobileNet~\cite{howard2017mobilenets}.
To solve~\eq{eq:full-goal} one may perform the following process:
\begin{enumerate}
    \item Find the largest $\omega$ such that $\calF(\omega\cdot \outdepth_{1:M}^\circ) \le \zeta$.
    \item Return $\omega\cdot \outdepth_{1:M}^\circ$.
\end{enumerate}
In most cases the form of $\calF$ allows for
easily finding the optimal $\omega$.
Thus, unlike other methods which require training
a network to determine which components are more or less
necessary, application of a width multiplier
is essentially free.
Despite its simplicity,
in 
our evaluations we found this approach to often give
good solutions, especially when $\outdepth_{1:M}^\circ$
is already a well-structured network.
The approach suffers, however, with decreased quality of the initial
network design.

Consider now an alternative, more sophisticated approach
based on sparsifying regularizers.
We may augment the objective~\eqref{eq:loss} with a regularizer $\calG(\theta)$
which induces sparsity in the neurons, putting greater cost on neurons which
contribute more to $\calF(\outdepth_{1:M})$.
The trained parameters $\theta^* = \text{argmin}_{\theta}~\{\calL(\theta) + \lambda\calG(\theta)\}$
then induce a new set of output widths $\outdepth_{1:M}'$
which are a tradeoff between optimizing the loss
given by $\calL$ and satisfying the constraint given by $\calF$.
Unlike the width multiplier approach,
this approach is able to change the relative sizes
of layers. However, the resulting structure $\outdepth_{1:M}'$
is not guaranteed to satisfy $\calF(\outdepth_{1:M}')\le \zeta$.
Moreover, this procedure often disproportionately 
sacrifices performance, especially when $\calF(\outdepth_{1:M}') <  \zeta$.

\subsection{Our Approach} \label{sec:our-approach}

We propose to utilize 
a hybrid of the two approaches, 
iteratively alternating between
a sparsifying regularizer and a 
uniform width multiplier.
Given a suitable regularizer $\calG$ 
which induces sparsity in the activations,
putting greater cost on activations which
contribute more to $\calF(\outdepth_{1:M})$
(we elaborate on the specific form of $\calG$
in subsequent sections),
we propose to
approximately solve~\eq{eq:full-goal}
starting from the seed network $\outdepth_{1:M}^\circ$ 
using Algorithm~\ref{algo:main}.

The \FlexNet algorithm optimizes the \DNN by iteratively shrinking (Steps 1-2) and expanding (usually, Step 3) 
the \DNN. 
At the shrinking stage, we apply a sparsifying regularizer on neurons.
This results in a \DNN that consumes less of the targeted resource, 
but typically achieves a lower performance. 
However, a key observation is that the training process in Step 1 not only highlights 
which layers of the \DNN are over-parameterized,
but also which layers are bottlenecked. 
For example, when targeting FLOPs, higher-resolution neurons in the lower layers of the \DNN tend to be sacrificed more than lower-resolution neurons in the upper layers of the \DNN. The situation is the exact opposite when the targeted resource is model size rather than FLOPs.

\begin{algorithm}[ht]
\caption{The \FlexNet Algorithm}
\begin{algorithmic}[1]
    \State Train the network to find
    \Statex $\theta^* = \argmin{\theta} \{\calL(\theta) + \lambda \calG(\theta)\}$,
    for suitable $\lambda$.
    \State Find the new widths $\outdepth_{1:M}'$ induced by $\theta^*$.
    \State Find the largests $\omega$ such that $\calF(\omega \cdot \outdepth_{1:M}') \le  \zeta$.
    \State Repeat from Step 1 for as many times as desired, setting $\outdepth_{1:M}^\circ = \omega\cdot\outdepth_{1:M}'$.\\
    \Return $\omega\cdot \outdepth_{1:M}'$.
\end{algorithmic}
\label{algo:main}
\end{algorithm} 

This leads us to Step 3 of the \FlexNet algorithm,
which usually performs an expansion. 
In this paper we only report one method for expansion, namely uniformly expanding all layer sizes via a width multiplier
as much as the constrained resource allows, 
although one may replace this with an alternative expansion technique.

We have thus completed one cycle of improving the network architecture, and we can continue this process iteratively until the performance is satisfactory, or until the \DNN architecture has converged (\ie further iterations lead to a near-identical \DNN structure). 
In our evaluation below, we found a single iteration of Steps 1-3
to be enough to yield a noticeable improvement over the na{\"i}ve
technique of just using a uniform width multiplier, while subsequent iterations
can bring additional benefits in performance. The optimal number of iterations, and whether the process converges, is yet to be investigated.
Note that a single iteration of the \FlexNet algorithm
comes at the cost of a number of training runs
equal to the number of values of $\lambda$ attempted,
often a small constant number (\ie 5 or less). 
Empirically, we found it easy to find a good range of $\lambda$ by trial-and-error.  Whether a value is too large or too small is evident very early on in training by observing if the constrained quantity collapses to zero or does not decrease at all.

We use the remainder of this section to elaborate
on the specifics of \FlexNet.
We begin by describing the calculation of $\calF$
for the two constraints we consider (FLOPs and model size).
We then describe how a penalty on
this constraint may be relaxed
to a simple yet surprisingly effective
regularizer $\calG$ with informative
sub-gradients.
Subsequently, we describe how to maintain
the sparsifying nature of $\calG$
when network topologies are not confined
to the traditional paradigm of stacked
layers with only local connections 
(\ie as in Residual Networks).
Extensions to \FlexNet
to make it applicable to design choices beyond
just layer widths are briefly discussed in the supplementary material

\subsection{Constraints}
In this paper we restrict the discussion to two simple types of constraints: the number of FLOPs per inference, and the model size (\ie number of parameters). However, our approach lends itself to generalizations to other constraints, provided that they can be modeled. 

Both the FLOPs and model size are dominated by layers associated with matrix multiplications - \ie convolutions. The FLOPs and model size are bilinear in the number of inputs and outputs of that layer:
\begin{equation}\label{bilinear}
\calF(\text{layer }L) = C(\inwidth_L,\inheight_L,\outwidth_L,\outheight_L,\filtwidth_L,\filtheight_L) \cdot \indepth_L \outdepth_L.
\end{equation}
In the case of a FLOPs constraint we have,
\begin{equation}
    C(\inwidth,\inheight,\outwidth,\outheight,\filtwidth,\filtheight) = 
    2 \outwidth\outheight\filtwidth\filtheight, \label{eq:flop_reg}
\end{equation}
and in the case of a model size constraint we have,
\begin{equation}
    C(\inwidth,\inheight,\outwidth,\outheight,\filtwidth,\filtheight) = 
    \filtwidth\filtheight. \label{eq:size_reg}
\end{equation}
For ease of notation, we will henceforth drop the arguments from $C$ and assume them to be implicit. The constraints also include the relatively small cost of the biases, which is linear in $\outdepth_L$, and omitted here to avoid clutter.

A sparsifying regularizer on neurons
will induce some of the neurons to be zeroed out.
Namely, the weight matrix will exhibit structured sparsity in such a way that the pre-activation at some index $i$ is zero for any input and the post-activation at the same
index is a constant.
Such neurons should be discounted from~\eq{bilinear}
since an equivalent network may be constructed without
the weights leading into and out of these neurons.
To reflect this, we rewrite~\eq{bilinear} as,
\begin{equation}\label{bilinear_a}
\calF(\text{layer }L) = C \sum_{i=0}^{\indepth_L-1} \inactive_{L,i} \sum_{j=0}^{\outdepth_L-1} \outactive_{L,j},
\end{equation}
where $\inactive_{L,i}$ ($\outactive_{L,j}$) is an indicator function which equals one if the $i$-th input ($j$-th output) of layer $L$ is {\em alive} -- not zeroed out. \eq{bilinear_a} represents an expression for the constrained quantity pertaining to a single convolution layer. The total constrained quantity is obtained by summing \eq{bilinear_a} over all layers in the \DNN:
\begin{equation}\label{full-constraint}
\calF(\outdepth_{1:M}) =\sum_{L=1}^{M+1} \calF(\text{layer }L).
\end{equation}

\subsection{Regularization}\label{sec:reg}
When shrinking a network, we wish to minimize the loss of the \DNN $\calL(\theta)$ subject to a constraint $\calF(\outdepth_{1:M}) \le \zeta$. 
The optimization problem is equivalent to applying
a penalty on the loss,
\begin{equation}
    \min_\theta \calL(\theta) + \lambda\calF(\outdepth_{1:M}),
\end{equation}
for a suitable $\lambda$.  
Note that $\calF$ is implicitly
a function of $\theta$, since its calculation (\eq{bilinear_a} and~\eq{full-constraint}) relies on indicator functions.
For tractable learning via gradient descent, 
it is necessary to replace the discontinuous L0 norm 
that appears in~\eq{bilinear_a} with a continuous proxy norm. There are many possible choices for this continuous proxy norm. 

In this work we choose to use an L1 norm on the 
$\gamma_L$ variables of batch normalization~\citep{batchnorm}.
We chose this regularization because it is
simple and widely applicable.
Indeed, many top-performing feed-forward models
apply batch normalization to each layer.
This means that each neuron has a particular $\gamma$
associated with it which determines its scale.  
Setting this $\gamma$ to zero
effectively zeros out the neuron.

Thus our relaxation of~\eq{bilinear_a} is
\begin{multline}\label{bilinear_reg}
    \calG(\theta, \text{layer }L) =  C \sum_{i=0}^{\indepth_L-1} |\gamma_{L-1,i}| \sum_{j=0}^{\outdepth_L-1} \outactive_{L,j} + \\ 
    C \sum_{i=0}^{\indepth_L-1} \inactive_{L,i} \sum_{j=0}^{\outdepth_L-1} |\gamma_{L,j}|,
\end{multline}
where for ease of notation 
we assume the input neurons to layer $L$ 
are given by layer $L-1$.
The regularizer for the whole network is then
\begin{equation}
\calG(\theta) =\sum_{L=1}^{M+1} \calG(\theta, \text{layer }L).
\end{equation}

Note that the $A$ and $B$ coefficients in~\eq{bilinear_reg} are dynamic quantities,
being piece-wise constant functions of the network weights. 
As neurons at the input of layer $L$ are zeroed out, 
the cost of each neuron at the output is reduced, and
vice versa for neurons at the output of layer $L$.  \eq{bilinear_reg} captures this behavior. 
In particular, \eq{bilinear_reg} is
discontinuous with respect to the $\gamma$'s. 
However, \eq{bilinear_reg} is still differentiable almost
everywhere, and thus we found that standard 
minibatch optimizers readily handle the discontinuity
of $\calG$.

\begin{figure}[t]
\vspace{-0.1cm}
\begin{center}
   \includegraphics[width=0.9\linewidth]{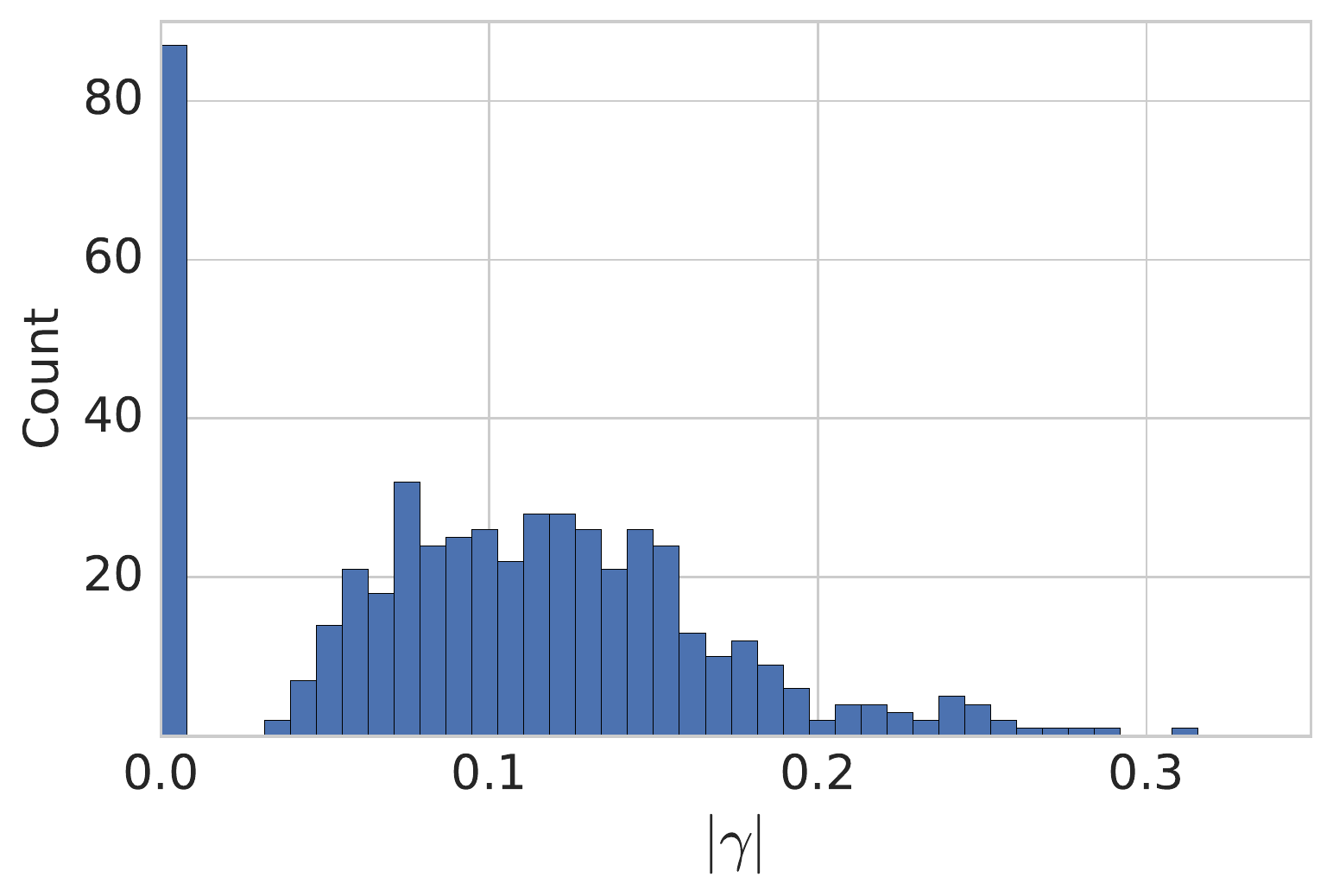}
\end{center}
\vspace*{-3mm}
   \caption{A histogram of $\gamma$ for one of the ResNet101 bottleneck layers when trained with a FLOP regularizer. Some of the $|\gamma|$'s are zeroed out, and are separated by a clear gap from the nonzero $|\gamma|$'s.}
\label{fig:gammas}
\end{figure}

While our regularizer is simple and general,
we found it to be surprisingly effective at inducing 
sparsity.
We show the induced values of $\gamma$ for one
network trained with $\calG$ in~\figref{fig:gammas}.
There is a clear separation between those $\gamma$'s
which have been zeroed out and those which continue
to contribute to the network's computation.

\comment{
    \subsection{Learning the network structure}
    As we show in Section ?? below, the $A$ (\eq{threshold}) of each neuron converges often times much sooner than the performance of the network does. The network has chosen which neurons to keep, and after a certain point no neurons die out or come back to life anymore. At this point the structure of the \DNN has converged. 
    
    At this point one can extract the learned number of outputs of each convolution, obtaining thereby a more lightweight (in the sense of consumption of the constrained resource) version of the initial \DNN. Alternatively, one could proceed to the next iteration: Assuming that  $\lambda$ was chosen such that the consumption of the constrained resource is below its target value, one can multiply the output sizes of all convolution by a factor that will bring it back to the target value or above it, and apply a second iteration of regularized training, according to \eq{eq:loss}. This process can be iterated until convergence. To test each learned \DNN structure obtained in the process, it has to be retrained without $\cal R$ to convergence.
}

\subsection{Preserving the Network Topology}
\DNNs in computer vision applications often have residual (skip) connections: \ie the input of layer $L_3$ can be the sum of the outputs of $L_1$ and $L_2$. If the outputs of $L_1$ and $L_2$ are regularized separately, it is not guaranteed that the exact same outputs will be zeroed out in $L_1$ and $L_2$, which can change the topology of the network and introduce new types of connectivity that did not exist before. 
While the latter is a legitimate modification of the network structure, it may result in a significant complication in the network structure when the network has tens of layers tied 
to each other via residual connections. 
To avoid these changes in the network topology, 
we group all neurons that are tied in skip connections via
a Group LASSO. 
For example, in the example above the $j$-th output of $L_1$ will be grouped with the $j$-th output of $L_2$. 
There are multiple ways to group them, and in the results presented in this work we use the $L_\infty$ norm - the maximum of the $|\gamma|$'s in the group.

\comment{
    \subsection{Network Expansion}
    So far we have described 
    a regularization which induces
    shrinking of a network.
    It is natural to extend \FlexNet
    to expanding the size of a network.
    Indeed, when observing the resulting
    $\gamma_L$'s of a network trained
    with our regularization it is not only
    clear which layers are over-parameterized
    (those layers which have a significant amount of
    zero-gammas) 
    but also which layers are bottle-necked
    (those layers which have very few
    or no zero-gammas).
    If one wishes to expand a network,
    then these bottle-necked layers 
    are where one should focus the expansion.
    
    The parameters of an
    expanded layer may be initialized 
    in a number of ways.
    Previous work has found that 
    random initialization 
    as linear combinations of existing 
    layers can work well~\citep{net2net}.
    In our experimental evaluation below,
    we opted to go for a much simpler method.
    We train the network once with
    the gamma-loss until convergence.
    We subsequently observe the 
    $\gamma_L$ values and produce a new
    network structure, 
    shrunk accordingly.
    Then we apply a uniform multiplicative
    expansion and retrain the network
    from scratch without the gamma loss.
    In this way, over-parameterized layers
    are shrunk while bottle-necked layers
    are expanded.  Furthermore, the 
    final performance of the network
    has no chance being adversely affected
    by residual effects of the L1-regularization.
}
 \secspace\section{Empirical Evaluation}
\label{sec:empirical}

We evaluate the \FlexNet algorithm for automatic structure learning
on a variety of datasets and seed network designs.
We give a brief overview of each experimental setup
in Section~\ref{sec:datasets}.
In Section~\ref{sec:case-study}, 
we go through in detail the application of \FlexNet on
one of these setups (Inception V2 on ImageNet),
examining the benefit and improvement at each step of the algorithm.
We then give a summarized view of the results of \FlexNet
applied to all datasets and all models in Section~\ref{sec:results}.
Finally, we take a closer look at our regularization in
Section~\ref{sec:regularizer}, 
showing that it adequately targets the desired constraint
using both quantitative and qualitative analysis.

\subsection{Datasets}
\label{sec:datasets}
We evaluate on a number of different
datasets encompassing various scales and domains.
\secspace\subsubsection{ImageNet}\secspace
ImageNet~\citep{ImageNet} is a well-known benchmark 
consisting of 1M images classified into 1000 distinct classes.
We apply \FlexNet on two markedly different 
seed architectures: 
Inception V2~\cite{inceptionv2}, and~\mobilenet~\cite{howard2017mobilenets}.
These two networks were the result of 
hand-tuning to achieve two distinct goals. 
The former network was designed to have maximal accuracy (on ImageNet) while the latter was designed to have low computation foot-print (\FLOPs) on mobile devices while maintaining good overall ImageNet accuracy.

For \mobilenet we use the smallest published resolution ($128\times128$) and the two smallest width multipliers ($50\%$ and $25\%$).  We choose these as it focuses \FlexNet 
on the low-\FLOPs regime, 
thus furthest away from the Inception V2 regime. 


\ignore{
}

\secspace\subsubsection{JFT}
At its introduction, ImageNet was significant
for its size.  
Recent years have seen ever larger datasets.
To evaluate the scalability of \FlexNet,
we choose the JFT dataset~\citep{distill,sun2017revisiting},
an especially large collection of labelled images, 
with about 350M images and about 20K labels. 
For this dataset we chose to start with the \resnet architecture \cite{he2016deep}, thus examining the applicability of \FlexNet to residual networks.

\secspace\subsubsection{AudioSet}
Finally, as a dataset encompassing a different
domain, we evaluate on AudioSet~\cite{audioset}.
The published AudioSet contains 
$2$M audio segments encompassing $500$ distinct labels.  
We use a larger version of the dataset
which contains $20$M labelled audio segments,
while maintaining approximately the same number of labels.
We seeded our model architecture with a residual network
based on a structure previously used 
for this dataset~\cite{hershey2017cnn}.

\subsection{A Case Study: Inception V2 on ImageNet}
\label{sec:case-study}
We provide a detailed look at each step of \FlexNet (described in Section~\ref{sec:our-approach})
on ImageNet with the seed network design $\outdepth_{1:M}^\circ$
corresponding to Inception V2~\cite{inceptionv2}.

\begin{figure}[ht]
\begin{center}
   \includegraphics[width=0.9\linewidth]{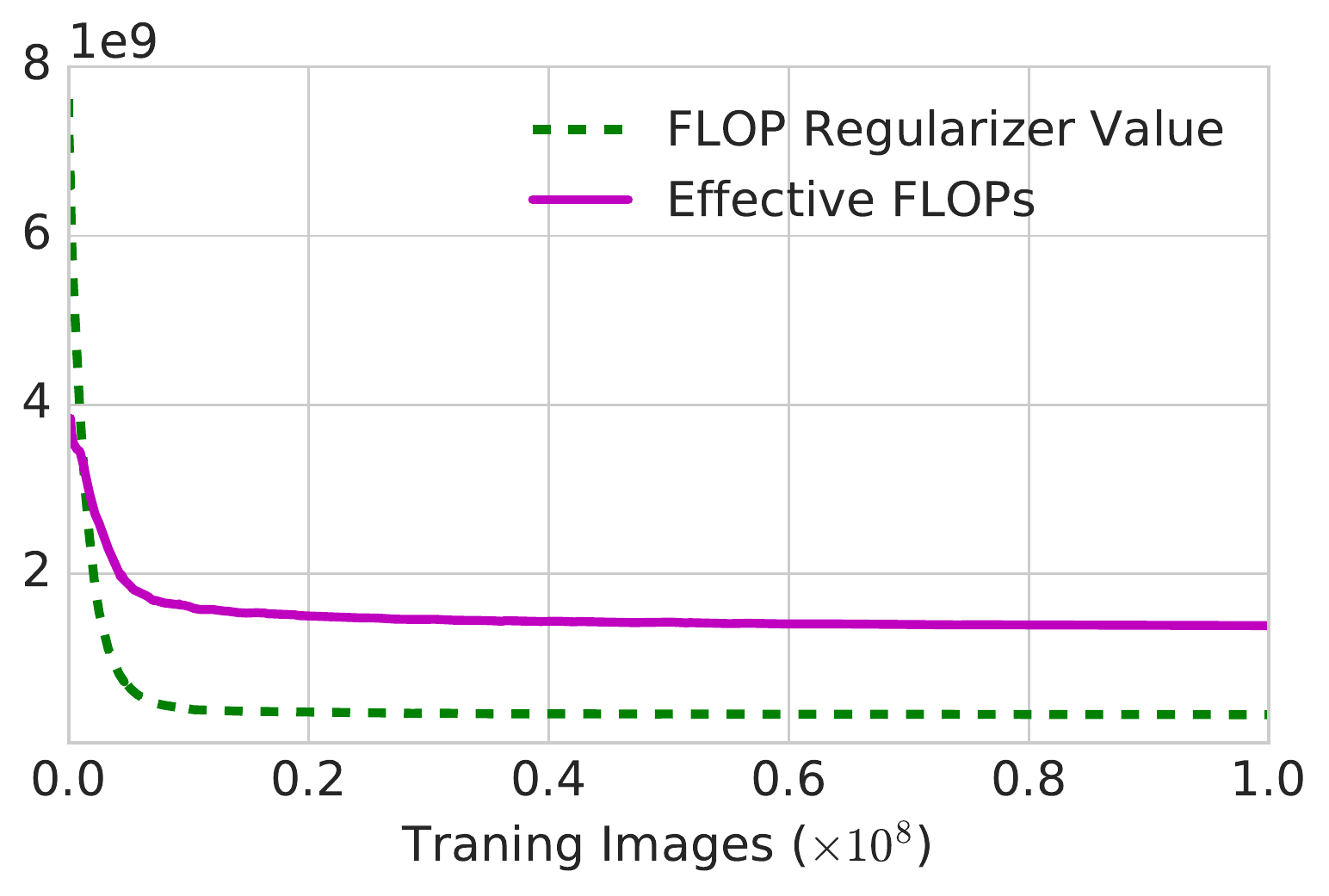}
\end{center}
   \caption{Rapid convergence of of the \FLOP\ regularization (green, dashed) and projected number of \FLOPs (purple) for ImageNet trained with a \FLOP\ regularizer strength of $\lambda=1.3\cdot10^{-9}$. 
   The projected number of \FLOPs 
   is computed by assuming all $|\gamma|<0.01$ are zeroed-out.
   }
\label{fig:convergence}
\end{figure}

The shrinking stage of \FlexNet trains 
the network with a sparsity-inducing
regularizer $\calG$.  
We use a \FLOPs-based regularizer and show the effect
of this regularizer on the actual \FLOPs during training
in~\figref{fig:convergence}.
Although the form of $\calG$ is only a proxy to the
true \FLOPs, it is clear that the regularizer
adequately targets the desired constraint.

\begin{figure}[ht]
\vspace{-0.2cm}
\begin{center}
   \includegraphics[width=1.0\linewidth]{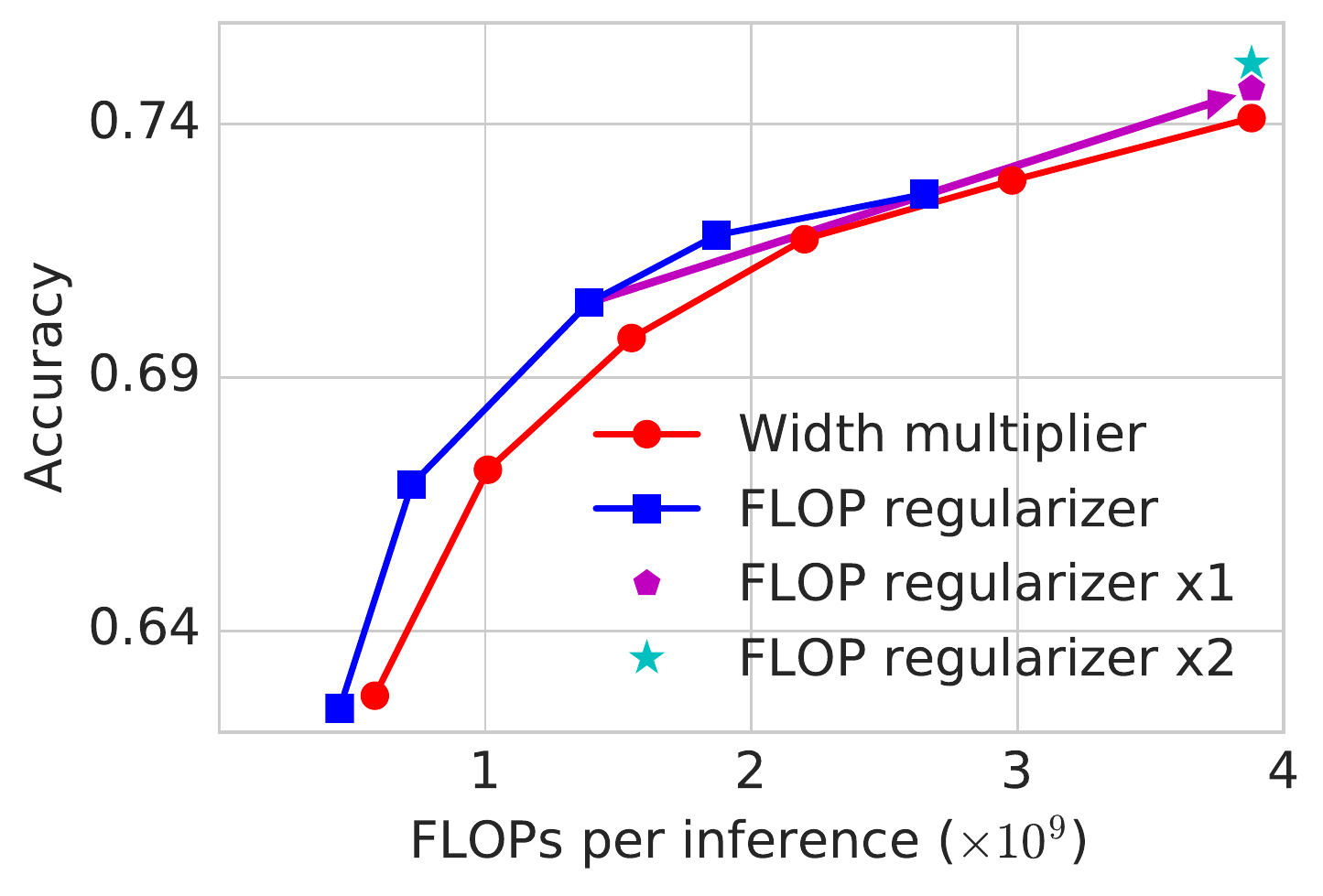}
\end{center}
   \vspace*{-3mm}
    \caption{
    ImageNet evaluation accuracy for various downsized versions of Inception V2
    using both a \naive width multiplier (red circles) and
    a sparsifying \FLOP\ regularizer (blue squares). 
    We also show the result of re-expanding
    one of the networks induced by the \FLOP\ regularizer
    to match the \FLOP\ cost of the original network (pentagon point).  
    A further increase in accuracy is achieved by performing the sparsifying and expanding process a second time (star point).
    }
\label{imagenet_fig}
\end{figure}

Applying $\calG$ with different strengths 
(different values of $\lambda$) leads to 
different shrunk networks.\footnote{{For a fixed $\lambda$, results are fairly reproducible across repeated experiments.  See the supplementary material.}}
We show the results of these distinct 
trained networks (blue line) compared
to a \naive application of the width multiplier (red line)
in~\figref{imagenet_fig}.
While it is clear that sparsifying using $\calG$
is more effective than applying a width multiplier,
our main goal in this work is to 
demonstrate that the accuracy of the \DNN can be 
\emph{improved} while maintaining 
a constrained resource usage (\FLOPs in this case).

This leads us to \FlexNet' expansion stage (Step 3).
We choose the \DNN obtained by using $\lambda=1.3\cdot 10^{-9}$ to 
re-scale using a uniform width multiplier until the number of \FLOPs per inference matches that of the seed Inception V2 architecture.
See results in~\figref{imagenet_fig} and Table~\ref{imagenet_table}.
The resulting \DNN achieves an improved accuracy compared to the Inception V2 baseline of 0.6\%. 
We then repeat our procedure again, 
first applying a sparsifying regularizer and then 
re-scaling to the original \FLOPs usage.
On the second iteration we achieve a further improvement of 0.5\%, adding up to a total improvement of 1.1\% compared to the baseline. 
Since the improved \DNN structures exhibited 
stronger overfitting than the seed, 
we introduced a dropout layer before the classifier
(crucially, we were not able to improve  
the accuracy of the seed network in a significant manner by applying dropout). 
The dropout values and the accuracies are summarized in 
Table \ref{imagenet_table}. 
Except for the dropout, all other hyperparameters used at training were identical for all \DNNs. 

In this case study we focused on improving accuracy while
preserving the FLOPs per inference.  However, it is clear that \FlexNet can trade-off the two objectives when a practitioner's priorities are different. 
For example, we found that the architecture learned in the second iteration can be shrunk by applying a width multiplier until the number of \FLOPs is reduced by 30\%, and the resulting \DNN matches the original Inception V2 accuracy.

\begin{table}[h]
\begin{center}
\begin{tabular}{|c|c|c|c|c| }
    \hline
Iteration   & $\omega$  & Dropout   & Weights               & Accuracy      \\ \hline
$0$           &   NA    & 0         &  $1.12\cdot 10^{7}$   & 74.1\% \\     
$1$           &  $1.69$ & 10\%      &  $1.61\cdot 10^{7}$   & 74.7\% \\     
$2$           &  $1.57$ & 20\%      &  $1.55\cdot 10^{7}$   & 75.2\% \\     
\hline
\end{tabular} 
\end{center}
\vspace*{-1mm}
\caption{
\FlexNet\ applied to the seed network of Inception V2 on ImageNet.
A regularization strength of $\lambda = 1.3\!\cdot\!10^{-9}$ was used in both iterations. The network was expanded to match the original \FLOPs of $3.88\!\cdot\!10^{9}$. Dropout rate was increased to mitigate over-fit caused by the increased model capacity.
Although the number of \FLOPs is constant, our method is capable of and chooses to increase the number of weights in the model.}
\label{imagenet_table}
\end{table}

\subsection{Improved Performance at No Cost}
\label{sec:results}
\begin{table}[h]
\begin{center}
\begin{tabular}{|l|c|c|c|}
\hline
Network & Baseline & \FlexNet & Relative Gain\\  
\hline\hline
 Inception V2      & 74.1     & 75.2 \ignore{74.7}  & +1.5\% \\
 \mobilenet $50\%$ & 57.1     & 58.1  & +1.78\%\\
 \mobilenet $25\%$ & 44.8     & 45.9  & +2.58\%\\
 \resnet           & 0.477    & 0.487  &+2.1\%\\
 AudioResNet       & 0.182     & 0.186  & +2.18\%\\
\hline
\end{tabular}
\end{center}
%
\caption{The result of applying \FlexNet to  a variety of datasets and model architectures while maintaining \FLOP\ cost.
\enote{removed:}
\ignore{Inception and \mobilenet are evaluated on ImageNet, \resnet on JFT, and AudioResNet on AudioSet. The reported results are for a single \FlexNet\ iteration (two for Inception V2), first shrinking the network via a sparsifying regularizer and then expanding to the original \FLOP\ cost via a uniform width multiplier.
Across all experimental evaluations,
we see increases in accuracy while using
the same number of \FLOPs.
}
}
\label{tab:results}
\end{table}

\begin{figure*}[h!]
  \begin{minipage}{\linewidth}
    \vspace{-0.2cm}
    \begin{center}
    \includegraphics[width=0.9\linewidth]{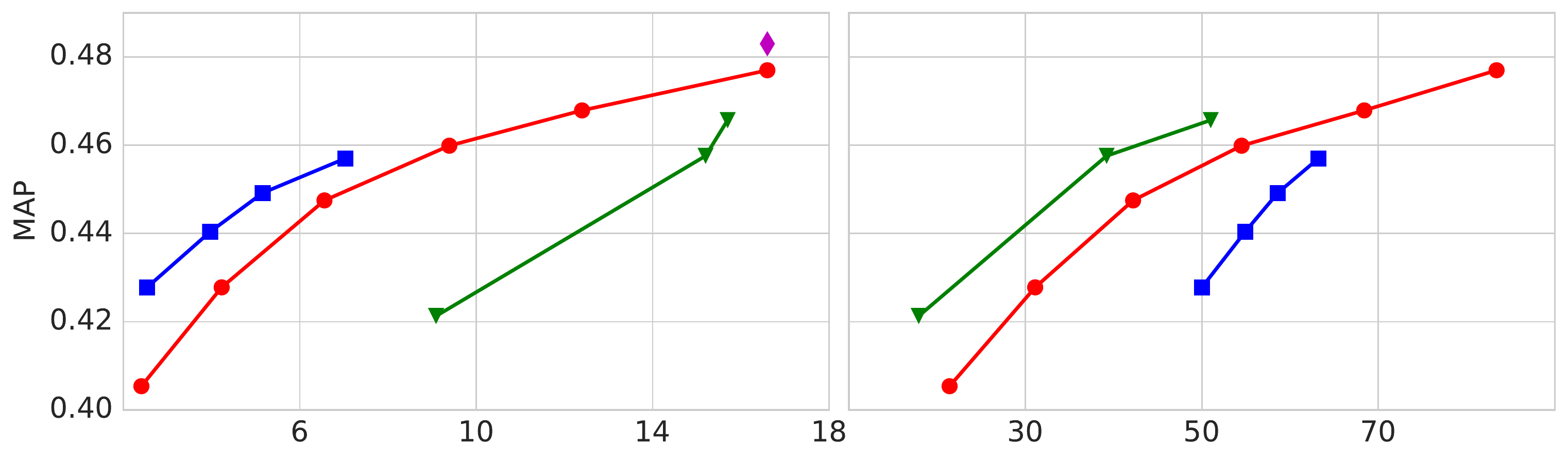}\hfill
    \end{center}
    \vspace{-0.75cm}
     \begin{center}
    \includegraphics[width=0.9\linewidth]{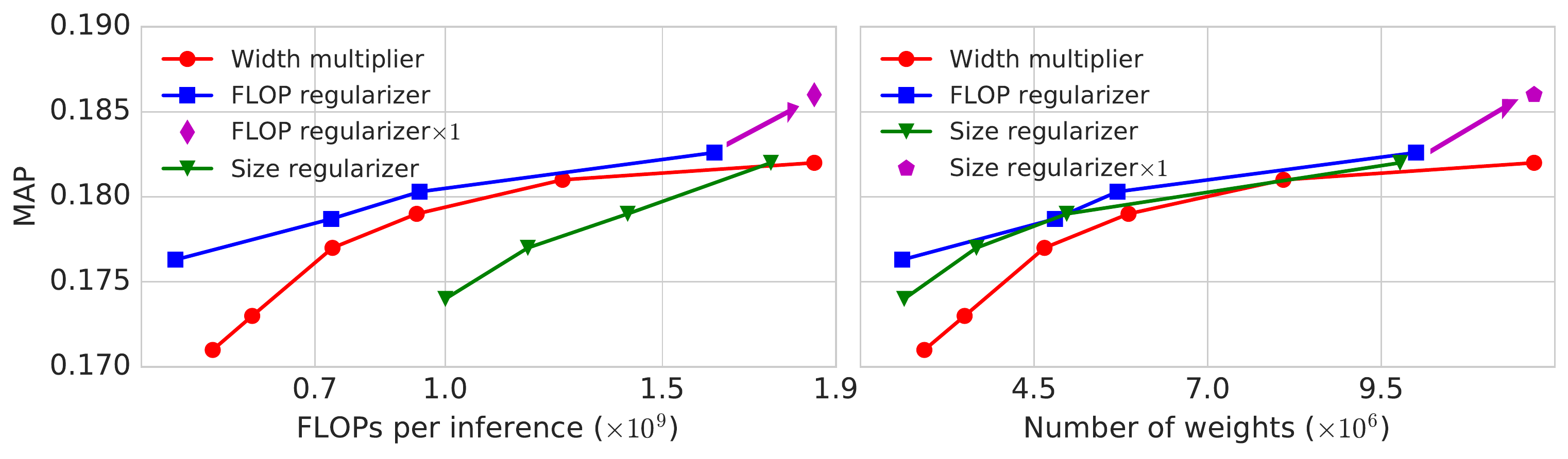} \hfill
    \end{center}
    \vspace{-0.4cm}
    \caption{
    MAP vs. \FLOPs (left) and 
    MAP vs. model-size (right) curves 
        on JFT (top) and 
        AudioSet (bottom).
    The magenta points in the AudioSet figures represent models expanded from a \FLOP\ (diamond) or size (pentagon) regularizes.  
    }\label{fig:mapvs}
\end{minipage}%

\vspace{-0.2cm}
\end{figure*}

We present the collective results of \FlexNet
on all experimental setups on a \FLOPs constraint
in Table~\ref{tab:results}.
In each setup we report the application
of \FlexNet to the seed network for a single iteration (two for Inception V2). Thus, each result requires up to three training
runs.

We see improvements in performance across
all datasets.
The 1\% improvement on \mobilenet is especially impressive because \mobilenet
was specifically hand-designed to optimize accuracy 
under a \FLOPs-constraint.

On JFT, an especially large dataset, we achieve over $2.1\%$ relative 
improvement.
We note that the first training run is run until the convergence
of the \FLOPs cost, which
is approximately 20 times faster than the convergence of the performance metric (MAP).
Thus, for a  given value of $\lambda$, a single iteration of \FlexNet adds only 5\% to the cost of training a single model. Since more than one attempt may be required to find a suitable $\lambda$, the actual added cost may be higher.

In AudioSet we continue to see the benefits of \FlexNet,
observing a 2.18\% relative increase in MAP.
To put this into perspective, an equivalent drop of 2.18\%
from the seed model corresponds to a \FLOPs per inference reduction of over 50\% (see~\figref{fig:mapvs}).

\subsection{Resource Targeting}
\label{sec:regularizer}
One of the contributions of this work is 
the form of the regularizer $\calG$, 
which methodically 
targets a particular resource. 
In this section we demonstrate its effectiveness. 

\figref{fig:mapvs} shows the results of applying a FLOPs-targeted $\calG$ and
a model size-targeted $\calG$ at varying strengths.
It is clear that the structures induced
when targeting FLOPs form a better 
FLOPs/performance tradeoff curve, but poor model size/performance tradeoff curves, and vice 
versa when targeting model size.

We may also examine the learned structures when targeting different resources. 
In~\figref{fig:structures} we present the induced
network structures when targeting FLOPs and when targeting model size.
One thing to notice is that the FLOP regularizer tends to remove neurons from the lower layers near the input, whereas the 
model size regularizer tends to remove neurons from upper layers near the output.
This makes sense, as the lower layers of the neural network are applied to a high-resolution image, and thus consume a large number of the total FLOPs.  In contrast, the upper layers of a neural network are typically 
where the number of channels is higher and thus contain larger weight matrices. The two very different learned structures in \figref{fig:structures} achieve similar MAP (0.428 and 0.421, whereas the baseline model with similar cost is 0.405).

An interesting byproduct of applying \FlexNet to residual networks is that the network also learns to \textit{shrink the number of layers}, as shown in the \FLOP\ regularized structure in \figref{fig:structures}. When all the residual filters in a layer are pruned, the output is a direct copy of the input and the layer essentially can be removed. Therefore \FlexNet achieves automatic layer shrinkage without any added complexity.

 \secspace\section{Conclusion}

We presented \FlexNet, a technique for learning \DNN structures under a constrained resource. 
In our analysis of FLOP and model size constraints,
we have shown that the form of the 
tradeoff between 
constraint and accuracy
is highly dependent on the specific resource, 
and that \FlexNet can successfully navigate
this tradeoff when targeting either FLOPs or model size. 
Furthermore, we have applied \FlexNet to
large scale problems to achieve improvements
over human-designed \DNN structures, with little extra training cost compared to training the \DNN once. While being highly effective, \FlexNet
is simple to implement and fast to apply,
and thus we hope it becomes a general
tool for machine learning practitioners
aiming to better automate the task of 
neural network architecture design.

 \section*{Acknowledgement}
We thank Mark Sandler, Sergey Ioffe, Anelia Angelova, and Kevin Murphy for fruitful discussions and comments on this manuscript.

{\small
\bibliographystyle{ieee}
\bibliography{main}

\begin{thebibliography}{10}\itemsep=-1pt

\bibitem{abadi2015tensorflow}
M.~Abadi, A.~Agarwal, P.~Barham, E.~Brevdo, Z.~Chen, C.~Citro, G.~S. Corrado,
  A.~Davis, J.~Dean, M.~Devin, et~al.
\newblock Tensorflow: Large-scale machine learning on heterogeneous systems,
  2015.
\newblock {\em Software available from tensorflow. org}, 1, 2015.

\bibitem{alvarez2016learning}
J.~M. Alvarez and M.~Salzmann.
\newblock Learning the number of neurons in deep networks.
\newblock In {\em Advances in Neural Information Processing Systems}, pages
  2270--2278, 2016.

\bibitem{chen2015net2net}
T.~Chen, I.~Goodfellow, and J.~Shlens.
\newblock Net2net: Accelerating learning via knowledge transfer.
\newblock {\em arXiv preprint arXiv:1511.05641}, 2015.

\bibitem{collins2014memory}
M.~D. Collins and P.~Kohli.
\newblock Memory bounded deep convolutional networks.
\newblock {\em arXiv preprint arXiv:1412.1442}, 2014.

\bibitem{ImageNet}
J.~Deng, W.~Dong, R.~Socher, L.-J. Li, K.~Li, and L.~Fei-Fei.
\newblock Imagenet: A large-scale hierarchical image database.
\newblock In {\em Computer Vision and Pattern Recognition, 2009. CVPR 2009.
  IEEE Conference on}, pages 248--255. IEEE, 2009.

\bibitem{denker1987large}
J.~Denker, D.~Schwartz, B.~Wittner, S.~Solla, R.~Howard, L.~Jackel, and
  J.~Hopfield.
\newblock Large automatic learning, rule extraction, and generalization.
\newblock {\em Complex systems}, 1(5):877--922, 1987.

\bibitem{feng2015learning}
J.~Feng and T.~Darrell.
\newblock Learning the structure of deep convolutional networks.
\newblock In {\em Proceedings of the IEEE International Conference on Computer
  Vision}, pages 2749--2757, 2015.

\bibitem{audioset}
J.~F. Gemmeke, D.~P.~W. Ellis, D.~Freedman, A.~Jansen, W.~Lawrence, R.~C.
  Moore, M.~Plakal, and M.~Ritter.
\newblock Audio set: An ontology and human-labeled dataset for audio events.
\newblock In {\em Proc. IEEE ICASSP 2017}, New Orleans, LA, 2017.

\bibitem{han2015learning}
S.~Han, J.~Pool, J.~Tran, and W.~Dally.
\newblock Learning both weights and connections for efficient neural network.
\newblock In {\em Advances in Neural Information Processing Systems}, pages
  1135--1143, 2015.

\bibitem{hassibi1993}
B.~Hassibi and D.~G. Stork.
\newblock Second order derivatives for network pruning: Optimal brain surgeon.
\newblock In {\em Advances in neural information processing systems}, pages
  164--171, 1993.

\bibitem{he2016deep}
K.~He, X.~Zhang, S.~Ren, and J.~Sun.
\newblock Deep residual learning for image recognition.
\newblock In {\em Proceedings of the IEEE conference on computer vision and
  pattern recognition}, pages 770--778, 2016.

\bibitem{hershey2017cnn}
S.~Hershey, S.~Chaudhuri, D.~P. Ellis, J.~F. Gemmeke, A.~Jansen, R.~C. Moore,
  M.~Plakal, D.~Platt, R.~A. Saurous, B.~Seybold, et~al.
\newblock Cnn architectures for large-scale audio classification.
\newblock In {\em Acoustics, Speech and Signal Processing (ICASSP), 2017 IEEE
  International Conference on}, pages 131--135. IEEE, 2017.

\bibitem{distill}
G.~Hinton, O.~Vinyals, and J.~Dean.
\newblock Distilling the knowledge in a neural network.
\newblock {\em arXiv preprint arXiv:1503.02531}, 2015.

\bibitem{howard2017mobilenets}
A.~G. Howard, M.~Zhu, B.~Chen, D.~Kalenichenko, W.~Wang, T.~Weyand,
  M.~Andreetto, and H.~Adam.
\newblock Mobilenets: Efficient convolutional neural networks for mobile vision
  applications.
\newblock {\em arXiv preprint arXiv:1704.04861}, 2017.

\bibitem{huang2017data}
Z.~Huang and N.~Wang.
\newblock Data-driven sparse structure selection for deep neural networks.
\newblock {\em arXiv preprint arXiv:1707.01213}, 2017.

\bibitem{batchnorm}
S.~Ioffe and C.~Szegedy.
\newblock Batch normalization: Accelerating deep network training by reducing
  internal covariate shift.
\newblock In {\em ICML}, pages 448--456, 2015.

\bibitem{nemo}
Y.-H. Kim, B.~Reddy, S.~Yun, and C.~Seo.
\newblock Nemo: Neuro-evolution with multiobjective optimization of deep neural
  network for speed and accuracy.

\bibitem{lebedev2016fast}
V.~Lebedev and V.~Lempitsky.
\newblock Fast convnets using group-wise brain damage.
\newblock In {\em Proceedings of the IEEE Conference on Computer Vision and
  Pattern Recognition}, pages 2554--2564, 2016.

\bibitem{leoptimal}
Y.~LeCun, J.~S. Denker, and S.~A. Solla.
\newblock Optimal brain damage.
\newblock In {\em Advances in neural information processing systems}, pages
  598--605, 1990.

\bibitem{liu2015sparse}
B.~Liu, M.~Wang, H.~Foroosh, M.~Tappen, and M.~Pensky.
\newblock Sparse convolutional neural networks.
\newblock In {\em Proceedings of the IEEE Conference on Computer Vision and
  Pattern Recognition}, pages 806--814, 2015.

\bibitem{liu2017learning}
Z.~Liu, J.~Li, Z.~Shen, G.~Huang, S.~Yan, and C.~Zhang.
\newblock Learning efficient convolutional networks through network slimming.
\newblock {\em arXiv preprint arXiv:1708.06519}, 2017.

\bibitem{molchanov2016pruning}
P.~Molchanov, S.~Tyree, T.~Karras, T.~Aila, and J.~Kautz.
\newblock Pruning convolutional neural networks for resource efficient transfer
  learning.
\newblock {\em arXiv preprint arXiv:1611.06440}, 2016.

\bibitem{murray2015auto}
K.~Murray and D.~Chiang.
\newblock Auto-sizing neural networks: With applications to n-gram language
  models.
\newblock {\em arXiv preprint arXiv:1508.05051}, 2015.

\bibitem{l2}
A.~Y. Ng.
\newblock Feature selection, l 1 vs. l 2 regularization, and rotational
  invariance.
\newblock In {\em Proceedings of the twenty-first international conference on
  Machine learning}, page~78. ACM, 2004.

\bibitem{real2017large}
E.~Real, S.~Moore, A.~Selle, S.~Saxena, Y.~L. Suematsu, Q.~Le, and A.~Kurakin.
\newblock Large-scale evolution of image classifiers.
\newblock {\em arXiv preprint arXiv:1703.01041}, 2017.

\bibitem{sandler2018inverted}
M.~Sandler, A.~Howard, M.~Zhu, A.~Zhmoginov, and L.-C. Chen.
\newblock Inverted residuals and linear bottlenecks: Mobile networks for
  classification, detection and segmentation.
\newblock In {\em Computer Vision and Pattern Recognition, 2018. CVPR 2018.
  IEEE Conference on}. IEEE, 2018.

\bibitem{sun2017revisiting}
C.~Sun, A.~Shrivastava, S.~Singh, and A.~Gupta.
\newblock Revisiting unreasonable effectiveness of data in deep learning era.
\newblock {\em arXiv preprint arXiv:1707.02968}, 1, 2017.

\bibitem{inceptionv2}
C.~Szegedy, V.~Vanhoucke, S.~Ioffe, J.~Shlens, and Z.~Wojna.
\newblock Rethinking the inception architecture for computer vision.
\newblock In {\em Proceedings of the IEEE Conference on Computer Vision and
  Pattern Recognition}, pages 2818--2826, 2016.

\bibitem{lasso}
R.~Tibshirani.
\newblock Regression shrinkage and selection via the lasso.
\newblock {\em Journal of the Royal Statistical Society. Series B
  (Methodological)}, pages 267--288, 1996.

\bibitem{tieleman2012lecture}
T.~Tieleman and G.~Hinton.
\newblock Lecture 6.5-rmsprop: Divide the gradient by a running average of its
  recent magnitude.
\newblock {\em COURSERA: Neural networks for machine learning}, 4(2):26--31,
  2012.

\bibitem{supernets}
T.~Veniat and L.~Denoyer.
\newblock Learning time-efficient deep architectures with budgeted super
  networks.
\newblock {\em arXiv preprint arXiv:1706.00046}, 2017.

\bibitem{wen2016learning}
W.~Wen, C.~Wu, Y.~Wang, Y.~Chen, and H.~Li.
\newblock Learning structured sparsity in deep neural networks.
\newblock In {\em Advances in Neural Information Processing Systems}, pages
  2074--2082, 2016.

\bibitem{williams1995bayesian}
P.~M. Williams.
\newblock Bayesian regularization and pruning using a laplace prior.
\newblock {\em Neural computation}, 7(1):117--143, 1995.

\bibitem{group-lasso}
M.~Yuan and Y.~Lin.
\newblock Model selection and estimation in regression with grouped variables.
\newblock {\em Journal of the Royal Statistical Society: Series B (Statistical
  Methodology)}, 68(1):49--67, 2006.

\bibitem{zhou2016less}
H.~Zhou, J.~M. Alvarez, and F.~Porikli.
\newblock Less is more: Towards compact cnns.
\newblock In {\em European Conference on Computer Vision}, pages 662--677.
  Springer, 2016.

\bibitem{zoph}
B.~Zoph and Q.~V. Le.
\newblock Neural architecture search with reinforcement learning.
\newblock {\em arXiv preprint arXiv:1611.01578}, 2016.

\bibitem{zoph2016neural}
B.~Zoph and Q.~V. Le.
\newblock Neural architecture search with reinforcement learning.
\newblock {\em arXiv preprint arXiv:1611.01578}, 2016.

\bibitem{nasnet}
B.~Zoph, V.~Vasudevan, J.~Shlens, and Q.~V. Le.
\newblock Learning transferable architectures for scalable image recognition.
\newblock {\em arXiv preprint arXiv:1707.07012}, 2017.

\end{thebibliography}
}

\twocolumn[]
\newpage 
\appendix
\newpage
\section{Inception V2 trained on ImageNet}
In this section we provide the technical details regarding the the experiments in Section 5.2 of the paper.

When training with a FLOP regularizer, we used a learning rate of $10^{-3}$, and we kept it constant in time. The values of $\lambda$ that were used to obtain the points displayed in Figure~4 are 0.7, 1.0, 1.3, 2.0 and 3.0, all times $10^{-9}$.

Tables \ref{imagenet_width_first} and \ref{imagenet_width_second} below lists the size of each convolution in Inception V2, for the seed network and for the two \FlexNet iterations. The names of the layers are the ones generated by
\href{https://github.com/tensorflow/tensorflow/blob/master/tensorflow/contrib/slim/python/slim/nets/inception_v2.py}{this}\footnote{\tiny{https://github.com/tensorflow/tensorflow/blob/master/tensorflow/contrib/slim/python/slim/nets/inception\_v2.py}} code. Each column represents a learned DNN structure, obtained from the previous one by applying a FLOP regularizer with $\lambda=1.3\cdot 10^{-9}$ and then the width multiplier that was needed to restore the number of FLOPs to the initial value of $3.88\cdot 10^9$. The width multipliers at iteration 1 and 2 respectively were 1.692 and 1.571.

\section{MobileNet Training Details}

\subsection{Training protocol}
Our models operate on $128\times128$ images. The training procedure is a slight variant of running the main \FlexNet algorithm for one iteration. This variability gives better results overall and is crucial for \FlexNet to overtake the $50\%$ width-multipler model (see below). The procedure is as follows:

\begin{enumerate}
    
\item The full network (width-multipler of $1.0$ on $128\times128$ image input) was first trained for $2$ million steps (which is the typical number of steps for a network's performance to plateau as observed from training models with similar model sizes). Note that training smaller networks (e.g. with a width-multiplier of $0.25$) takes significantly more steps, e.g. around $10$ millions steps, to converge.

\item The checkpoint was used to initialize \FlexNet training, which goes on for an additional $10$ million steps or until the FLOPs of the active channels converge, whichever is longer. We tried a range of $\lambda$ values $\in \{3, 4, \ldots, 10, 11\} \times 10^{-9}$ to ensure that the converged FLOPs remain close to the FLOPs of the width-multiplier baselines.
\item We took the converged checkpoint and extracted a pruned network (both structure and weights) that consists of only the active channels.

\item Finally, we fine-tuned the pruned network using a small learning rate ($0.0013$). This is merely to restore moving average statistics for batch-normalization, and normally takes a negligible number, e.g. $20k$, of steps. While training for longer keeps improving the accuracy, simply training for $20k$ steps suffices to outperform models with width multipliers.
\end{enumerate}

All training steps use the same optimizer, which is discussed below.

\subsection{Trainer} \label{sec:mobilenet-opt}
We use the same trainer from MobileNet v2~\cite{sandler2018inverted}, described below.
We trained with the RMSProp optimizer~\cite{tieleman2012lecture} implemented in Tensorflow~\cite{abadi2015tensorflow} with a batch-size of $96$. The initial learning rate was chosen from $\{0.013, 0.045\}$, unless otherwise specified. The learning rate decays by a factor of $0.98$ every $2.5$ epochs. Training uses $16$ workers asynchronously. 
\subsection{Observations}
\begin{figure}[t!]
\vspace{-0.2cm}
\begin{center}
   \includegraphics[width=1\linewidth]{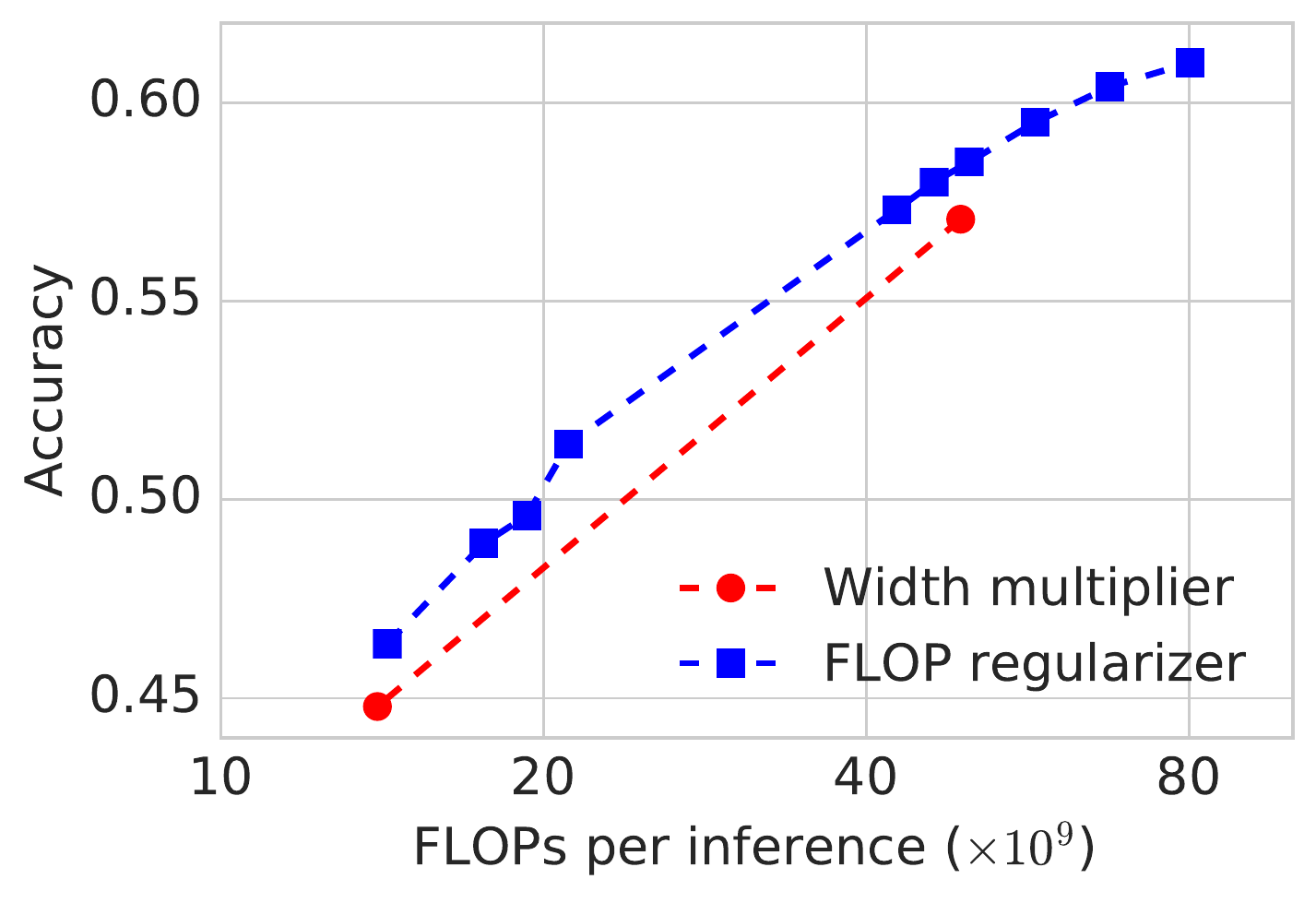}
\end{center}
   \vspace*{-3mm}
    \caption{
    ImageNet evaluation accuracy for various MobileNets on $128\times128$ images
    using both a \naive\,width multiplier (red circles) and
    a sparsifying \FLOP\ regularizer (blue squares). 
    }
\label{fig:mobilenet:sat}
\end{figure}

The total training time for each attempted $\lambda$ value is around $2+10=12$ million steps, which is less than twice the number of steps (around $10$ million) for training a regular network. Although multiple $\lambda$ values are required, each one of them contributes to the ``optimal'' FLOPs-vs-accuracy tradeoff, as shown in figure~\ref{fig:mobilenet:sat}. The ``optimality'' is defined in a narrow sense that no model is dominated in both FLOP and accuracy by another. By contrast, the $50\%$ width-multipler model is dominated by the \FlexNet models. Finally, we found that both the learning rate and the $\lambda$ parameter affects the converged FLOPs, but just the $\lambda$ parameter by itself suffices to traverse the range of desirable FLOPs.

\begin{figure*}[t!]
\vspace{-0.2cm}
\begin{center}
   \includegraphics[width=1.0\linewidth]{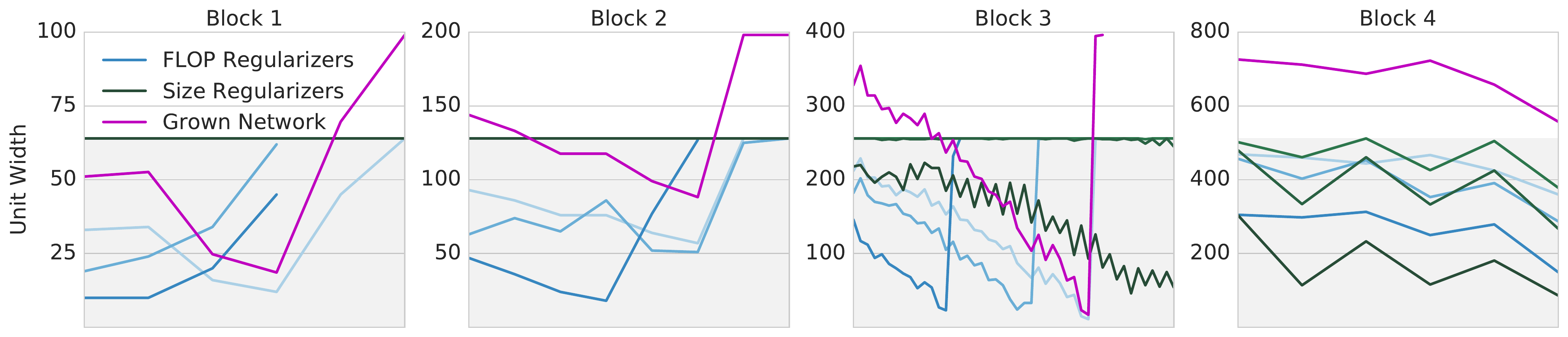}
\end{center}
   \vspace*{-3mm}
    \caption{
    Each of the four figures show the width of units in  \resnet\ blocks (1-4). The green (blue) shaded lines represent different strength of the FLOP (size) regularizer.
    The purple line represents the unit width of a model expanded from a FLOP-regularized \resnet\ model so that the number of \FLOPs matches these of the seed model. One can observe that increasing strengths of the FLOP regularizer (darker blue) remove more and more neurons from all blocks, and remove entire residual units (layers) from all blocks except for Block 4. Increasing the strength of the size regularizer (darker green) mainly removes neurons from Block 4.
    }
\label{fig:someresnets}
\end{figure*}

\begin{figure}[t!]
    \centering
    \includegraphics[width=1.0\linewidth]{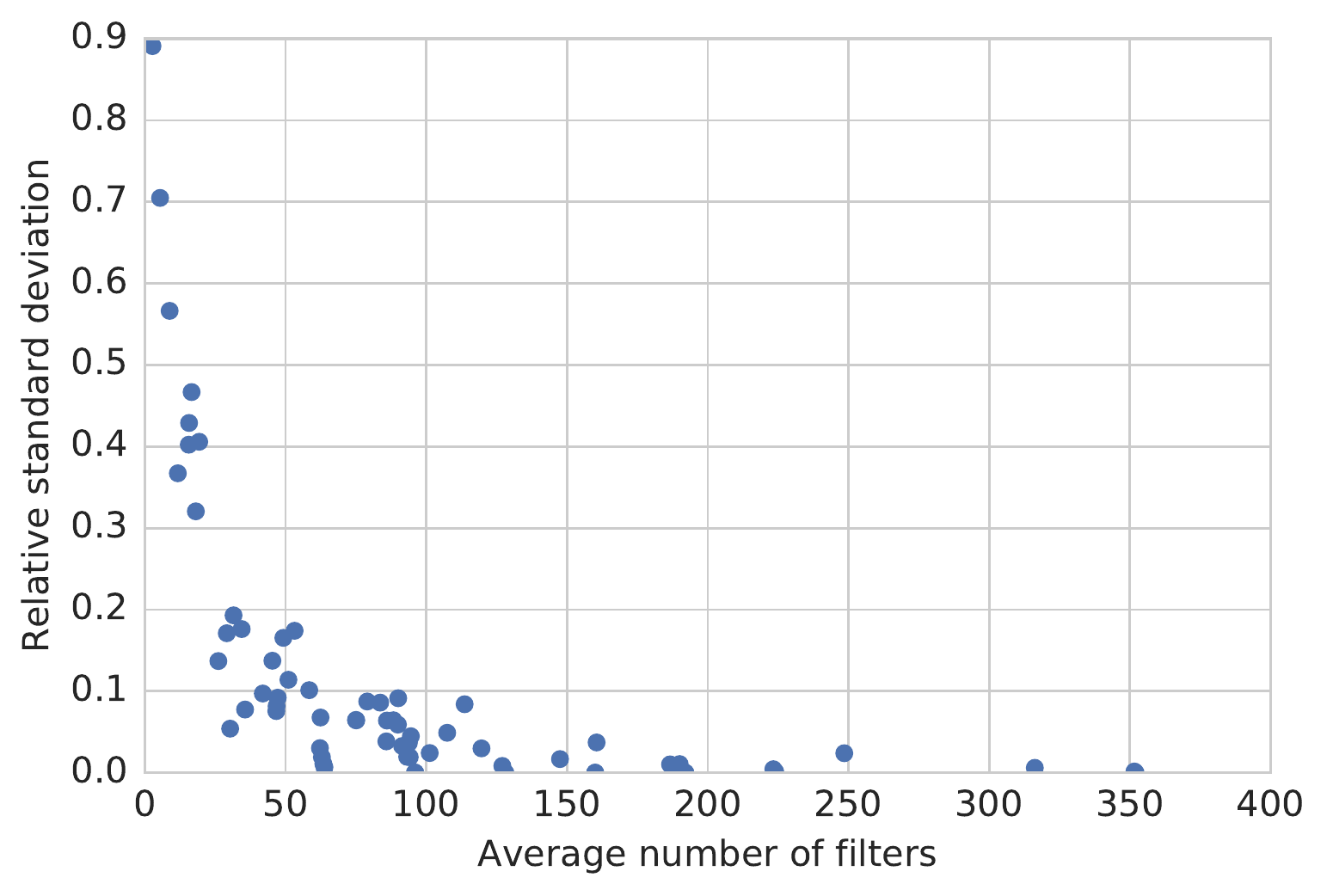}
    \caption{A scatter plot of relative standard deviations v.s.\ average number of filters of each layer in ImageNet Inception V2 model. The standard deviation was calculated over the results of 10 independent runs of Inception V2 with a FLOP regularizer of $\lambda=1.3\cdot 10^{-9}$, using the same hyperparameter configuration.}
    \label{fig:inception_layer_size_std}
\end{figure}

\section{\resnet\ on JFT}
The FLOP regularizer $\lambda$-s used in Figure 5 on JFT were 0.7, 1.0, 1.3 and 2 times $10^{-9}$. The size regularizer $\lambda$-s were 0.7, 1 and 3 times $10^{-7}$. The width multiplier values were 1.0, 0.875, 0.75, 0.625, 0.5, and 0.375. \figref{fig:someresnets} illustrates the structures learned when applying these regularizers on ResNet101.

\section{Stability of \FlexNet}

\begin{figure}[t!]
    \centering
    \includegraphics[width=1.0\linewidth]{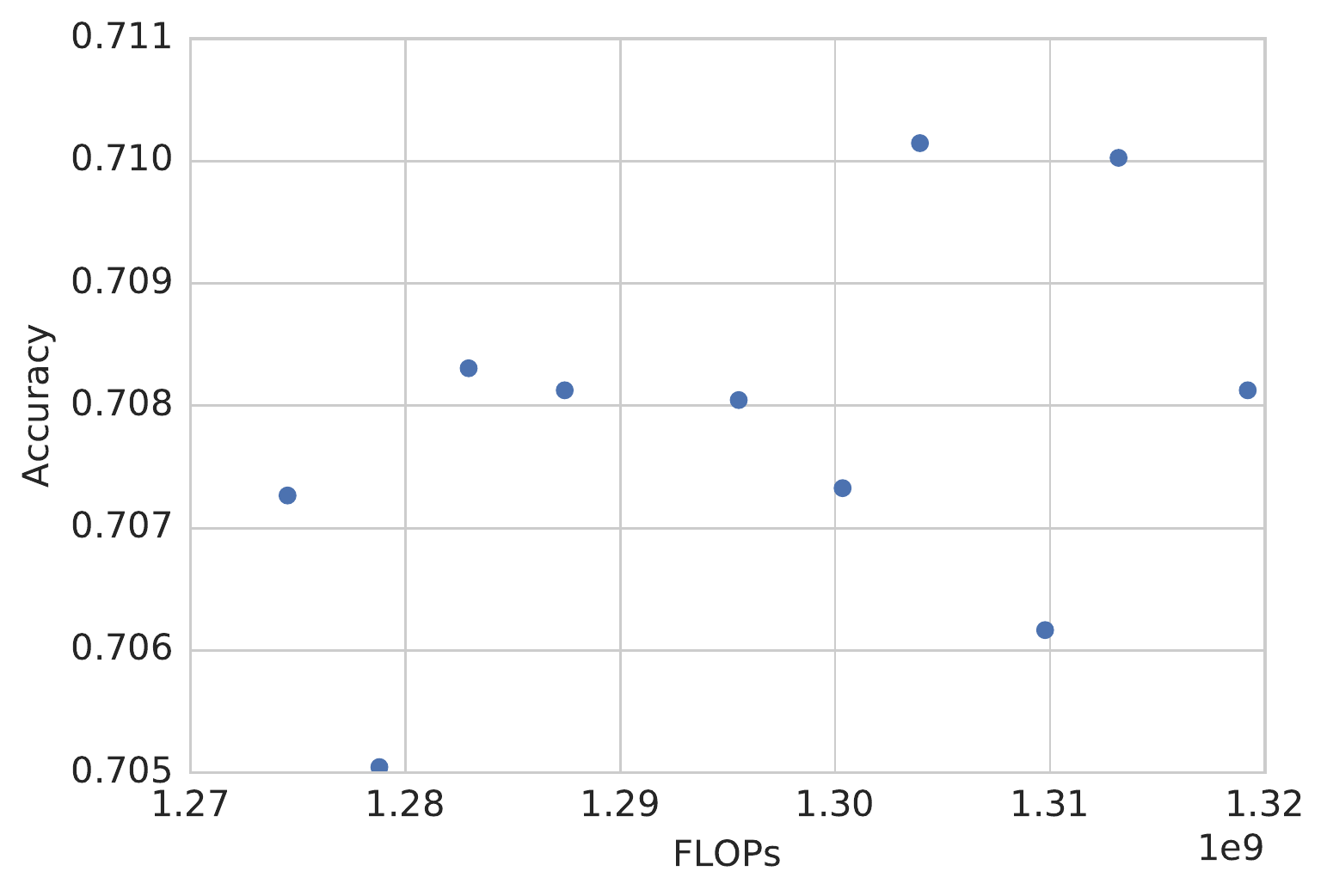}
    \caption{FLOPs v.s.\ test accuracy for Inception V2 model on the ImageNet dataset. Each point represents an independent run of Inception V2 with a FLOP regularizer of $\lambda=1.3\cdot 10^{-9}$, using the same hyperparameter configuration. The differences in the FLOP counts of the resulting architectures and in their test accuracies is shown in the figure. The relative
        standard deviation for FLOPs and test accuracy across 10 runs are \textbf{1.12\%} and \textbf{0.208\%} respectively.}
    \label{fig:inception_flops}
\end{figure}

In this section, we study the stability of \FlexNet with Inception V2 model on the ImageNet dataset. We trained the Inception V2 model
regularized by FLOP regularizer with a constant learning rate of $10^{-3}$. We also set the value of $\lambda$ to be $1.3 \times 10^{-9}$.
The training procedure was repeated independently for 10 times. We extracted the final architecture, e.g. the number of filters in each layer,
generated by \FlexNet from each run, and computed the relative standard deviations\footnote{Standard deviation divided by the mean.}
(RSTD) for the number of filters in each layer of the Inception V2 model across the 10 independent runs.
Figure~\ref{fig:inception_layer_size_std} shows the
scatter plot of RSTD for the ImageNet Inception V2 model. Such results show that the number of filters in most of the layers
does not change too much across different runs of \FlexNet with the same parameter configuration. Few of the layers have slightly
large RSTD. However the number of filters in these layers is small, which means the absolute changes of the number of filters in these
layers are still quite small across independent runs. Figure~\ref{fig:inception_flops} shows the scatter plot of FLOPs v.s.\ test accuracy
of Inception V2 model retrained over ImageNet dataset with the network architectures generated by the 10 independent runs of \FlexNet with
FLOPs regularizer. As we can see from this figure, the FLOPs and test accuracies from different runs all converged to the same region
with a relative standard deviation of \textbf{1.12\%} and \textbf{0.208\%} respectively, which are relatively small. All of these results
demonstrate that the \FlexNet is capable of generating pretty stable DNN architectures under constrained computation resources. 

\newpage

\begin{table*}[h]
\begin{center}
\begin{tabular}{|c|c|c|c|c| }
    \hline
    Layer name & iteration 0 & iteration 1 & iteration 2 \\
    \hline
Conv2d\_1a\_7x7 & 64 & 78 & 86 \\
Conv2d\_2b\_1x1 & 64 & 51 & 25 \\
Conv2d\_2c\_3x3 & 192 & 217 & 309 \\
Mixed\_3b/Branch\_0/Conv2d\_0a\_1x1 & 64 & 108 & 170 \\
Mixed\_3b/Branch\_1/Conv2d\_0a\_1x1 & 64 & 81 & 0 \\
Mixed\_3b/Branch\_1/Conv2d\_0b\_3x3 & 64 & 73 & 0 \\
Mixed\_3b/Branch\_2/Conv2d\_0a\_1x1 & 64 & 73 & 112 \\
Mixed\_3b/Branch\_2/Conv2d\_0b\_3x3 & 96 & 42 & 63 \\
Mixed\_3b/Branch\_2/Conv2d\_0c\_3x3 & 96 & 61 & 96 \\
Mixed\_3b/Branch\_3/Conv2d\_0b\_1x1 & 32 & 52 & 69 \\
Mixed\_3c/Branch\_0/Conv2d\_0a\_1x1 & 64 & 108 & 170 \\
Mixed\_3c/Branch\_1/Conv2d\_0a\_1x1 & 64 & 15 & 24 \\
Mixed\_3c/Branch\_1/Conv2d\_0b\_3x3 & 96 & 8 & 13 \\
Mixed\_3c/Branch\_2/Conv2d\_0a\_1x1 & 64 & 19 & 0 \\
Mixed\_3c/Branch\_2/Conv2d\_0b\_3x3 & 96 & 0 & 0 \\
Mixed\_3c/Branch\_2/Conv2d\_0c\_3x3 & 96 & 17 & 0 \\
Mixed\_3c/Branch\_3/Conv2d\_0b\_1x1 & 64 & 108 & 168 \\
Mixed\_4a/Branch\_0/Conv2d\_0a\_1x1 & 128 & 130 & 75 \\
Mixed\_4a/Branch\_0/Conv2d\_1a\_3x3 & 160 & 154 & 82 \\
Mixed\_4a/Branch\_1/Conv2d\_0a\_1x1 & 64 & 54 & 66 \\
Mixed\_4a/Branch\_1/Conv2d\_0b\_3x3 & 96 & 86 & 69 \\
Mixed\_4a/Branch\_1/Conv2d\_1a\_3x3 & 96 & 154 & 154 \\
Mixed\_4b/Branch\_0/Conv2d\_0a\_1x1 & 224 & 377 & 573 \\
Mixed\_4b/Branch\_1/Conv2d\_0a\_1x1 & 64 & 108 & 121 \\
Mixed\_4b/Branch\_1/Conv2d\_0b\_3x3 & 96 & 159 & 107 \\
Mixed\_4b/Branch\_2/Conv2d\_0a\_1x1 & 96 & 161 & 124 \\
Mixed\_4b/Branch\_2/Conv2d\_0b\_3x3 & 128 & 178 & 53 \\
Mixed\_4b/Branch\_2/Conv2d\_0c\_3x3 & 128 & 181 & 83 \\
Mixed\_4b/Branch\_3/Conv2d\_0b\_1x1 & 128 & 201 & 258 \\
Mixed\_4c/Branch\_0/Conv2d\_0a\_1x1 & 192 & 325 & 496 \\
Mixed\_4c/Branch\_1/Conv2d\_0a\_1x1 & 96 & 134 & 13 \\
Mixed\_4c/Branch\_1/Conv2d\_0b\_3x3 & 128 & 147 & 11 \\
Mixed\_4c/Branch\_2/Conv2d\_0a\_1x1 & 96 & 144 & 162 \\
Mixed\_4c/Branch\_2/Conv2d\_0b\_3x3 & 128 & 154 & 118 \\
\hline
\end{tabular} 
\end{center}
\vspace*{-1mm}
\caption{}
\label{imagenet_width_first}
\end{table*}

\begin{table*}[h]
\begin{center}
\begin{tabular}{|c|c|c|c|c| }
    \hline
    Layer name & iteration 0 & iteration 1 & iteration 2 \\
    \hline
Mixed\_4c/Branch\_2/Conv2d\_0c\_3x3 & 128 & 135 & 146 \\
Mixed\_4c/Branch\_3/Conv2d\_0b\_1x1 & 128 & 217 & 303 \\
Mixed\_4d/Branch\_0/Conv2d\_0a\_1x1 & 160 & 271 & 424 \\
Mixed\_4d/Branch\_1/Conv2d\_0a\_1x1 & 128 & 105 & 94 \\
Mixed\_4d/Branch\_1/Conv2d\_0b\_3x3 & 160 & 118 & 90 \\
Mixed\_4d/Branch\_2/Conv2d\_0a\_1x1 & 128 & 51 & 80 \\
Mixed\_4d/Branch\_2/Conv2d\_0b\_3x3 & 160 & 39 & 61 \\
Mixed\_4d/Branch\_2/Conv2d\_0c\_3x3 & 160 & 58 & 91 \\
Mixed\_4d/Branch\_3/Conv2d\_0b\_1x1 & 96 & 162 & 255 \\
Mixed\_4e/Branch\_0/Conv2d\_0a\_1x1 & 96 & 162 & 255 \\
Mixed\_4e/Branch\_1/Conv2d\_0a\_1x1 & 128 & 110 & 64 \\
Mixed\_4e/Branch\_1/Conv2d\_0b\_3x3 & 192 & 130 & 82 \\  
Mixed\_4e/Branch\_2/Conv2d\_0a\_1x1 & 160 & 32 & 50 \\
Mixed\_4e/Branch\_2/Conv2d\_0b\_3x3 & 192 & 22 & 35 \\
Mixed\_4e/Branch\_2/Conv2d\_0c\_3x3 & 192 & 36 & 57 \\
Mixed\_4e/Branch\_3/Conv2d\_0b\_1x1 & 96 & 162 & 255 \\
Mixed\_5a/Branch\_0/Conv2d\_0a\_1x1 & 128 & 217 & 324 \\
Mixed\_5a/Branch\_0/Conv2d\_1a\_3x3 & 192 & 325 & 482 \\
Mixed\_5a/Branch\_1/Conv2d\_0a\_1x1 & 192 & 151 & 237 \\
Mixed\_5a/Branch\_1/Conv2d\_0b\_3x3 & 256 & 73 & 113 \\
Mixed\_5a/Branch\_1/Conv2d\_1a\_3x3 & 256 & 404 & 635 \\
Mixed\_5b/Branch\_0/Conv2d\_0a\_1x1 & 352 & 596 & 936 \\
Mixed\_5b/Branch\_1/Conv2d\_0a\_1x1 & 192 & 321 & 11 \\
Mixed\_5b/Branch\_1/Conv2d\_0b\_3x3 & 320 & 535 & 17 \\
Mixed\_5b/Branch\_2/Conv2d\_0a\_1x1 & 160 & 271 & 258 \\
Mixed\_5b/Branch\_2/Conv2d\_0b\_3x3 & 224 & 379 & 178 \\
Mixed\_5b/Branch\_2/Conv2d\_0c\_3x3 & 224 & 379 & 200 \\
Mixed\_5b/Branch\_3/Conv2d\_0b\_1x1 & 128 & 217 & 341 \\
Mixed\_5c/Branch\_0/Conv2d\_0a\_1x1 & 352 & 596 & 930 \\
Mixed\_5c/Branch\_1/Conv2d\_0a\_1x1 & 192 & 257 & 102 \\
Mixed\_5c/Branch\_1/Conv2d\_0b\_3x3 & 320 & 168 & 110 \\
Mixed\_5c/Branch\_2/Conv2d\_0a\_1x1 & 192 & 313 & 300 \\
Mixed\_5c/Branch\_2/Conv2d\_0b\_3x3 & 224 & 272 & 146 \\
Mixed\_5c/Branch\_2/Conv2d\_0c\_3x3 & 224 & 178 & 226 \\
Mixed\_5c/Branch\_3/Conv2d\_0b\_1x1 & 128 & 217 & 341 \\  
\hline
\end{tabular} 
\end{center}
\vspace*{-1mm}
\caption{}
\label{imagenet_width_second}
\end{table*}

\section{Extensions of the method}
\label{sec:extensions}
We have restricted the discussion and evaluation in this paper
to optimizing only the output widths $\outdepth_{1:M}$ 
of all layers.
However, our iterative process 
of shrinking via a sparsifying regularizer
and expanding via a uniform multiplicative factor
easily lends itself to optimizing over other 
aspects of network design.

For example, to determine filter dimensions
and network depth, previous work~\citep{wen2016learning} 
has proposed
to leverage Group LASSO and residual connections
to induce structured sparsity corresponding
to smaller filter dimensions and reduced network depth.
This gives us a suitable shrinking mechanism.
For expansion, one may reuse the idea of the width multiplier
to uniformly expand all filter dimensions and network depth.
To avoid a substantially larger network, it may be beneficial
to incorporate some rules regarding which filters
will be uniformly expanded 
(\eg by observing which filters were least affected
by the sparsifying regularizer; or more simply by random
selection).

\end{document}